\documentclass[10pt,twocolumn,letterpaper]{article}

\usepackage[pagenumbers]{cvpr} 

\usepackage[T1]{fontenc}    
\usepackage{microtype}      
\usepackage{cuted}         
\usepackage{currfile}      
\usepackage{tgadventor}     

\usepackage{array}          
\usepackage{nicefrac}       
\usepackage{amsfonts}       
\usepackage{url}            
\usepackage{wrapfig}        
\usepackage[most]{tcolorbox}
\tcbuselibrary{skins,breakable}

\definecolor{ghcolor}{rgb}{0.15,0.15,0.18}
\definecolor{webcolor}{rgb}{0.31,0.45,0.62}

\definecolor{cvprblue}{rgb}{0.21,0.49,0.74}
\usepackage[pagebackref,breaklinks,colorlinks,allcolors=cvprblue]{hyperref}

\def\confName{CVPR}
\def\confYear{2027}

\title{SignVerse-2M: A Two-Million-Clip Pose-Native Universe \\of 55+ Sign Languages}

\author{%
Sen Fang$^{1}$\quad Hongbin Zhong$^{2}$\quad Yanxin Zhang$^{3}$\quad Liwei Che$^{1}$\quad Dimitris N.~Metaxas$^{1}$\\
$^{1}$Rutgers University\quad
$^{2}$Georgia Institute of Technology\quad
$^{3}$University of Wisconsin--Madison\\
\href{https://signerx.github.io/SignVerse-2M}{\textbf{{\color{magenta}\nolinkurl{https://signerx.github.io/SignVerse-2M}}}}\\[-20pt]
}

\begin{document}
\maketitle
\begin{strip}
\centering
\includegraphics[width=\linewidth]{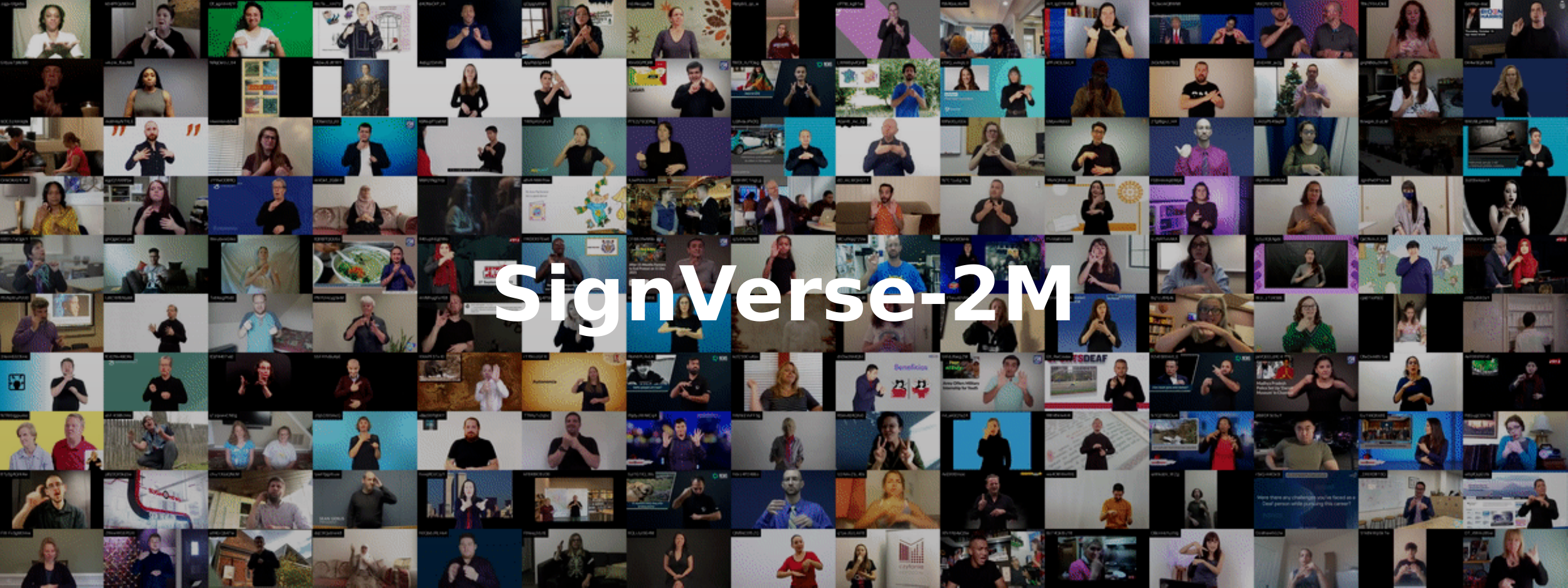}
\captionsetup{hypcap=false}
\captionof{figure}{\textbf{SignVerse-2M at scale.} The corpus spans diverse signers, languages, clothing, backgrounds, viewpoints, and recording conditions from open-world sign language videos. We convert this visual diversity into a unified pose-native interface for multilingual modeling and modern pose-driven generation.}
\captionsetup{hypcap=true}
\label{fig:teaser}%
\end{strip}

\begin{abstract}
Large-scale sign language resources usually pair RGB videos with text and support recognition, translation, and generation. However, RGB frames contain backgrounds, clothing, and other appearance information unrelated to signing motion. Representing the videos as whole-body keypoints such as DWPose removes much of this variation, produces a compact description of body, hand, and facial movement, and matches the input used by recent pose-guided video models. Existing datasets do not provide this representation at scale for in-the-wild sign language videos.

We present \textbf{SignVerse-2M}, a multilingual dataset containing \textbf{approximately two million pose clips from more than 55 sign languages}. We process publicly available sign language videos with a shared DWPose pipeline, representing body, hand, and facial motion in the same keypoint format across all clips. The source videos include diverse signers, viewpoints, and recording environments. We also release the processing pipeline, task definitions, and SignDW Transformer baselines. Experiments show that the data supports multilingual text-to-pose modeling and that the generated poses can directly drive several recent pose-conditioned rendering systems.
\end{abstract}

\section{Introduction}

\begin{figure*}[t]
    \centering
    \includegraphics[width=.99\textwidth]{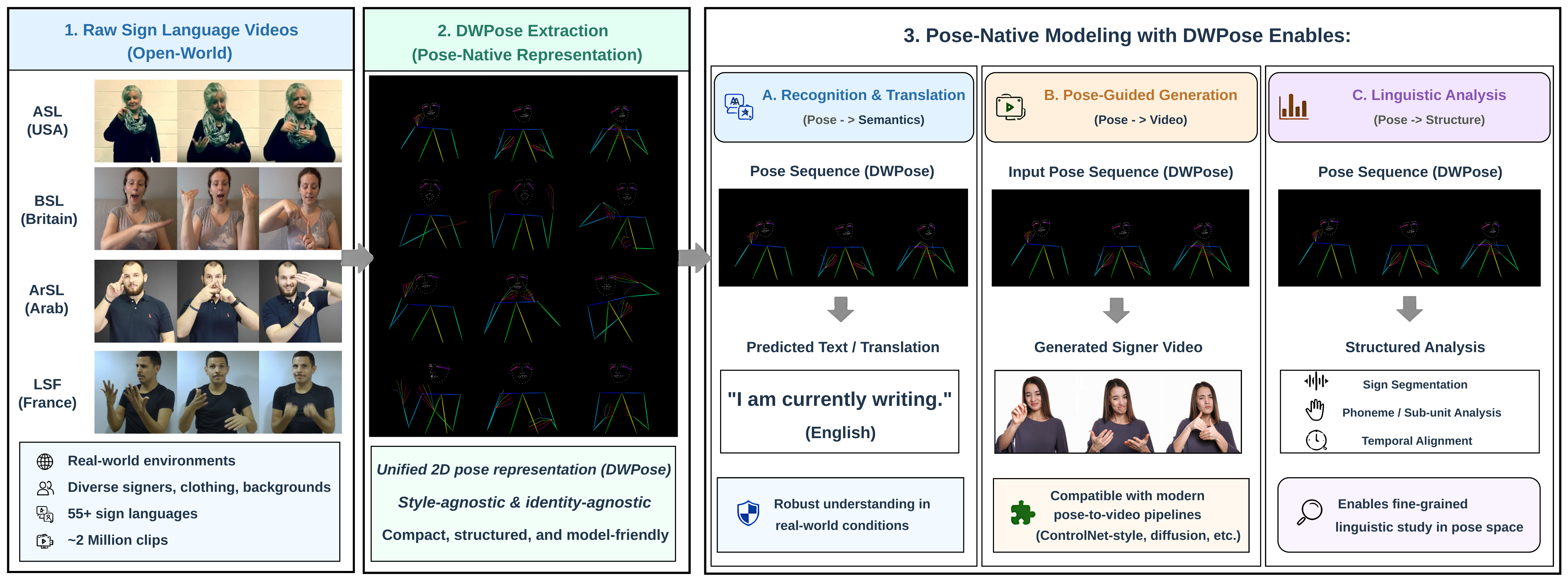}
    \caption{\textbf{Overview of SignVerse-2M.} SignVerse-2M organizes large-scale public sign language videos into a unified pose-native interface for multilingual sign language modeling. This representation is designed to be directly consumable by modern pose-driven generation pipelines and can serve as an intermediate control space for digital human or avatar generation \cite{chen2023executing,cai2023smplerx,zwitserlood2004synthetic,zhang2023adding}.}
    \label{fig:cover}
\end{figure*}

In recent years, the scale and coverage of sign language datasets have grown substantially with the expansion of online video resources and multimodal learning methods \cite{Duarte_2021_how2sign,camgoz2018neural,camgoz2020sign,forster2012rwth,MS-ASL,WLASL}. However, existing large-scale sign language resources still face two main limitations: first, they are mostly designed for tasks such as sign language recognition, retrieval, or sign-to-text translation \cite{hu2023continuous,skeleton_aware_slr,slt-how2sign-wicv2023,camgoz2020sign,koller2020quantitative}, with supervision centered on weak or strong alignment between videos and text \cite{camgoz2018neural,Duarte_2021_how2sign,albanie2020bsl1k}; second, they often involve only a small number of signers recorded with fixed clothing and backgrounds, in environments that are neither dynamic nor diverse and are overly idealized \cite{forster2012rwth,RWTH-PHOENIX-Weather-2014,SIGNUM,Bohacek_2022_WACV}. These resources are valuable for understanding what is being expressed in sign language videos, but they do not naturally provide a unified representation and evaluation interface for open-environment recognition or for generation conditioned on complex textual inputs \cite{fang2026signxcontinuoussignrecognition,Fang_2025_ICCV,fang2025signdiffdiffusionmodelamerican}. This issue is especially pronounced in multilingual sign language settings, where raw videos contain strong nuisance factors such as background, viewpoint, signer identity, and recording conditions, while different datasets often adopt \textbf{incompatible representations, preprocessing pipelines, and evaluation protocols}, making model comparison and interpretation of conclusions difficult \cite{yin2022mlslt,JAsigning,moryossef2021datasets}.

This leads us to believe that what sign language research currently lacks is not simply more videos, but a \textbf{unified, pose-driven evaluation resource for in-the-wild sign language tasks} \cite{saunders2020progressive,Fang_2025_ICCV,baltatzis2024neuralsignactorsdiffusion,stoll2020text2sign}. This need is particularly evident in sign language generation: existing models based on Mesh Human, MediaPipe, or non-standard OpenPose derivatives \cite{cai2023smplerx,MediaPipe,cao2018openpose} are often cumbersome to connect with contemporary pose-to-video pipelines \cite{ma2024followposeposeguidedtexttovideo,zhang2025mimicmotionhighqualityhumanmotion,cheng2025wan}.

Over the past three years, a large number of pose-driven human image and video generation models have gradually converged on a unified technical interface: they first extract standardized keypoints such as DWPose~\cite{yang2023effective} from real-world videos, and then use these keypoints to drive ControlNet-style control modules \cite{zhang2023adding,peng2025controlnextpowerfulefficientcontrol} or other pose-conditioned generation frameworks \cite{zhang2023controlvideotrainingfreecontrollabletexttovideo,chan2019everybodydance}. In other words, in the broader visual generation community, DWPose~\cite{yang2023effective}, as a high-accuracy and high-speed 2D skeleton estimation model, has effectively become one of the \textbf{standard input representations for modern pose-driven generation}.

\begin{table*}[t]
\centering
\resizebox{0.99\textwidth}{!}{
\begin{tabular}{lccccl}
\toprule
Dataset                                                                               & Duration(h) & Vocabulary(k) & Annotation Type               & Year  & Domain                 \\ \midrule
KETI~\cite{Ko-2019-SLT-based-human-keypoint-estimation}                              & 27.99      & 0.49          & Spoken Text                   & 2019  & Emergency situations   \\
PHOENIX-2014T~\cite{camgoz2018neural}                           & 10.50      & 3.90          & Spoken Text, Gloss            & 2018  & Weather Forecast       \\
CSL Daily~\cite{Zhou2021ImprovingSLT-with-monolingual-CSLDaily}                      & 23.27      & 4.60          & Spoken Text, Gloss            & 2021  & Daily life             \\
OpenASL~\cite{shi-etal-2022-openASL}                                                 & 288        & 33            & Spoken Text                   & 2022  & Youtube (news + vlogs) \\
How2Sign~\cite{Duarte_2021_how2sign}                                                 & 79.10      & 22.40         & Spoken Text                   & 2021  & Instructional          \\
ASLLRP~\cite{neidle2022asl}                                                          & 80         & 2.1           & Gloss                         & 2022  & Comprehensive          \\
YouTube-ASL~\cite{uthus2023youtubeasl}                                               & $\sim$1000 & 60            & Spoken Text                   & 2023  & Youtube (open-domain ASL) \\
\midrule
YouTube-SL-25~\cite{tanzer2024youtubesl25}                                           & 3207       & $\sim$374     & Spoken Text                   & 2024  & Youtube (55+ languages) \\
\textbf{SignVerse-2M (ours)}                                                         & 3207       & $\sim$374     & Spoken Text, Pose             & 2026  & Pose-native open-world multilingual \\
\bottomrule
\end{tabular}
}
\vspace{4pt}
\caption{\textbf{
Comparison with representative and widely used sign language datasets:
} We compare SignVerse-2M with representative and widely used sign language datasets under a unified layout, covering annotation type, scale, and domain. The top five rows mainly correspond to spoken-text translation benchmarks, while ASLLRP, YouTube-ASL, and YouTube-SL-25 place our resource in broader and more open-domain settings. Duration and vocabulary are reported as corpus-level statistics; ``-'' denotes values that are unavailable or not directly comparable, and approximate values are marked with ``$\sim$''.}
\label{tab:SLT_datasets}
\vspace{-12pt}
\end{table*}


Motivated by this perspective, we propose \textbf{SignVerse-2M}. We build on existing large-scale multilingual sign language video resources \cite{tanzer2024youtubesl25} and apply DWPose for unified pose extraction and standardized processing, converting them into a large-scale pose-native sign language representation dataset covering more than 55 sign languages and about two million clips. As shown in Figure~\ref{fig:cover}, our goal is not to recreate yet another video-text corpus, but to construct a data interface that can directly support sign language generation research: 1. it reduces appearance variation and shifts the focus back to the hand, body, and upper-body motions themselves; 2. it provides a consistent input-output space for cross-lingual modeling, unified training, and standardized evaluation; 3. it is directly compatible with existing pose-to-pose, pose-to-video, and other pose-driven generation pipelines, making it easier to connect sign language generation research to mainstream pose-conditioned modeling paradigms. As shown in Table~\ref{tab:SLT_datasets}, representative existing datasets differ substantially in annotation type, scale, and domain, but most are still organized around video--text supervision rather than a unified pose-native interface.

To address these questions, this paper makes four main contributions:
\begin{itemize}
    \item We construct \textbf{SignVerse-2M}, a large-scale, multilingual, uniformly represented pose-native sign language dataset that systematically converts public video resources into a pose-sequence corpus suitable for generative tasks.
    \item We explicitly position this resource as a \textbf{sign language interface for modern pose-driven generative models}, emphasizing its compatibility with the DWPose-centered control paradigm and treating pose extraction as a core representation step.
    \item We introduce \textbf{task settings and a simple SignDW Transformer baseline} for sign language generation research to validate the practical usability of this resource and provide a reproducible starting point for future work.
    \item We release the processing pipeline, structured metadata, person-centric pose schema, visualization tools, predefined splits, and quality-tracking records to support inspection and reproducible reuse of the corpus.
\end{itemize}

In summary, SignVerse-2M is intended to support a shift in sign language research from idealized laboratory settings toward more comparable pose-based modeling on open-world data.

\section{Background and Positioning}

\subsection{Traditional Video-Text Sign Language Corpora}
\noindent\textbf{The Development of Sign Language Data.}~
Sign language resources have progressed from domain-specific corpora such as PHOENIX \cite{forster2012rwth,RWTH-PHOENIX-Weather-2014}, MS-ASL \cite{MS-ASL}, and How2Sign \cite{Duarte_2021_how2sign} to larger open-domain and multilingual collections, including WLASL \cite{WLASL}, BSL-1K \cite{albanie2020bsl1k}, BOBSL \cite{Albanie-2021-BOBSL}, OpenASL \cite{shi-etal-2022-openASL}, MLSLT \cite{yin2022mlslt}, YouTube-ASL \cite{uthus2023youtubeasl}, and YouTube-SL-25 \cite{tanzer2024youtubesl25}. These resources are primarily organized as video--text pairs for recognition, retrieval, and sign-to-text translation \cite{camgoz2018neural,camgoz2020sign,hu2023continuous,slt-how2sign-wicv2023,skeleton_aware_slr,koller2020quantitative}. Sign language production has developed in parallel through progressive pose generation \cite{saunders2020progressive}, SignLLM and its Prompt2Sign setting \cite{Fang_2025_ICCV}, diffusion-based production \cite{fang2025signdiffdiffusionmodelamerican,baltatzis2024neuralsignactorsdiffusion}, hierarchical rendering \cite{fang2025stablesignerhierarchicalsign,fang2026stablesignerv34d}, and autoregressive pose modeling \cite{yin2024t2sgptdynamicvectorquantization}. These systems make representation and evaluation compatibility increasingly important \cite{stoll2020text2sign,saunders2020progressive,walsh2025slrtp2025signlanguageproduction}. Raw RGB remains valuable, but it also carries background, signer appearance, viewpoint, and recording conditions into modeling and evaluation \cite{Bohacek_2022_WACV}.

\begin{figure*}[t]
    \centering
    \includegraphics[width=\textwidth]{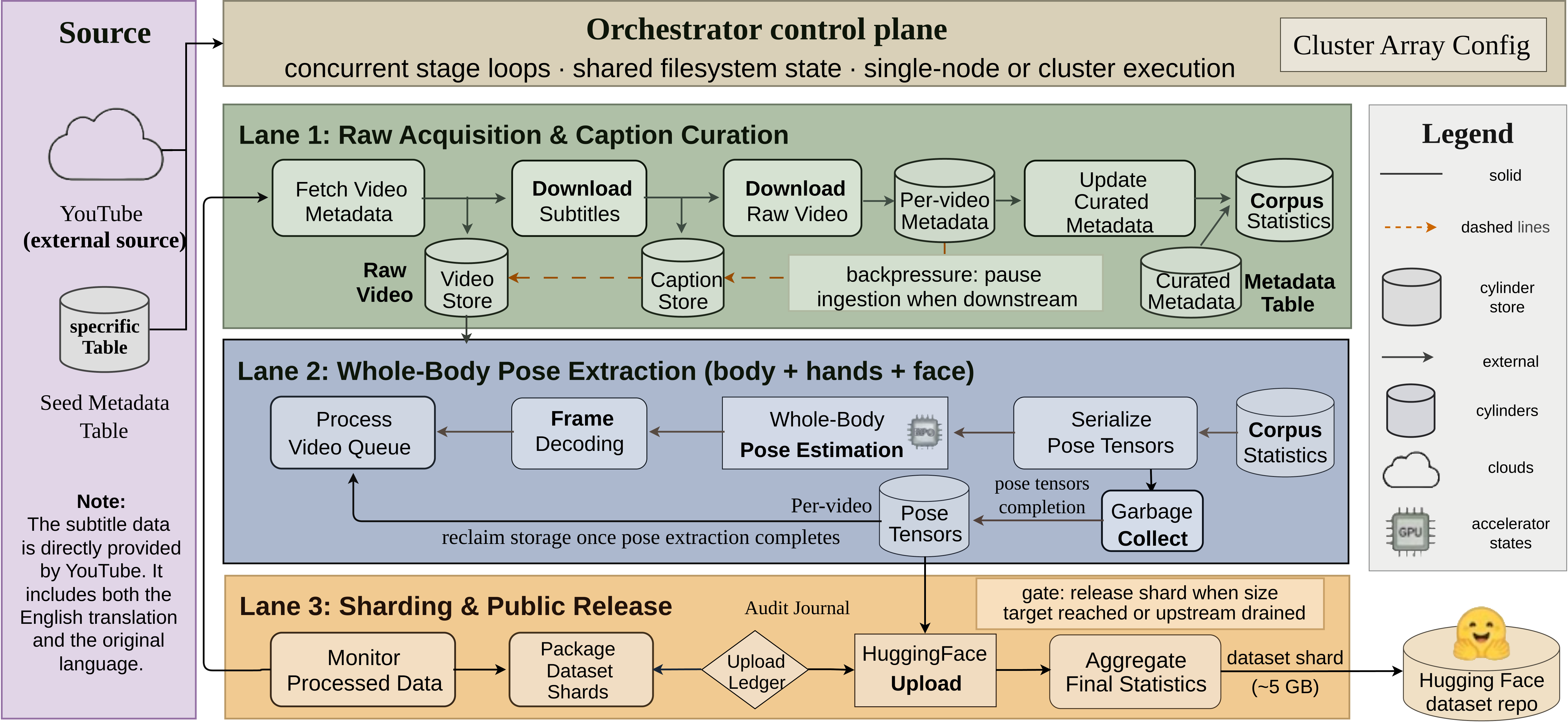}
    \caption{\textbf{Overview of the SignVerse-2M data processing pipeline.} The pipeline is organized into three lanes: raw acquisition and caption curation, whole-body pose extraction, and sharding for public release. Starting from a manifest indexed by `video\_id` and `sign\_language`, the system retrieves metadata, subtitles, and raw videos, converts each video into DWPose-based body, hand, and face keypoint sequences, and then packages the outputs into shards for publication. The orchestration layer maintains status records, supports staged re-execution and failure recovery, and enables the corpus to be built incrementally rather than as a one-shot preprocessing export.}
    \label{fig:signverse_data_pipeline}
\end{figure*}

\subsection{From Laboratory Distributions to Real-World Distributions}
\noindent\textbf{Why Choose a Pose-Native Representation?}~
Pose representation provides a more suitable intermediate space for this problem. Recent LVLM analyses show that information flow and shallow-layer visual attention affect how models select visual evidence \cite{zhang-etal-2025-redundancy,zhang-etal-2025-shallow}, while fine-grained visual reasoning also depends on individuating and aggregating local entities \cite{che2026mllmsstrugglecountovercoming}. These questions are relevant to signing, where body, hands, and face provide distinct but coordinated evidence. Multimodal learning under heterogeneous or incomplete modality combinations further motivates representations that remain consistent across varied inputs \cite{s23156986,10.1007/978-3-031-70378-2_25}. By uniformly mapping raw videos into keypoint sequences \cite{yang2023effective,cao2018openpose,MediaPipe,cai2023smplerx}, models can focus more directly on the temporal structure and body geometry of sign language motion without having to simultaneously handle complex appearance variation \cite{saunders2020progressive,saunders2021continuous,stoll2020text2sign}. This is especially important in multilingual settings, where visual conditions vary substantially across languages, video sources, and signers \cite{yin2022mlslt,Fang_2025_ICCV}. More importantly, over the past three years, DWPose \cite{yang2023effective} has become one of the standardized control interfaces for a large number of pose-driven image and video generation models \cite{ma2024followposeposeguidedtexttovideo,zhang2025mimicmotionhighqualityhumanmotion,cheng2025wan}. Whether in ControlNet-style conditional control \cite{zhang2023adding,zhang2023controlvideotrainingfreecontrollabletexttovideo,peng2025controlnextpowerfulefficientcontrol} or in other generation frameworks that use human keypoints as intermediate conditions \cite{chan2019everybodydance,saunders2020everybodysignnowtranslating}, a stable, general, and large-scale extractable pose representation is essential. A concrete example is that RGB-based recognition models are difficult to apply directly in real-world environments: once the background, person, or clothing changes, the input video features can differ significantly \cite{koller2020quantitative,Bohacek_2022_WACV}. Pose-based models suppress much of this appearance variation by construction and have supported both sign recognition and generation \cite{fang2026signxcontinuoussignrecognition,fang2025stablesignerhierarchicalsign}.

\subsection{Positioning of SignVerse-2M}
\noindent\textbf{Relation to Existing Multilingual Sign Language Resources.}~
SignVerse-2M inherits from public multilingual sign language video corpora \cite{Duarte_2021_how2sign,albanie2020bsl1k,Albanie-2021-BOBSL,shi-etal-2022-openASL,yin2022mlslt} and builds specifically on the scale and language coverage of YouTube-SL-25 \cite{tanzer2024youtubesl25}. Prior work primarily asks how to collect, align, and organize cross-lingual sign language video-text data from the web in order to support translation tasks \cite{camgoz2020sign,camgoz2018neural}. Recent generative research improves efficient flow generation \cite{fang2026streamflowtheoryalgorithmimplementation}, flow-based autoencoding \cite{fang2026racrectifiedflowauto}, and preference-conditioned image generation \cite{mo2025prefgenmultimodalpreferencelearning}. In parallel, pose-conditioned systems increasingly use standardized keypoints as control signals \cite{yang2023effective,cheng2025wan,zhang2023adding}. SignVerse-2M connects these developments by transforming public sign language videos into a unified pose space for reusable and comparable generative research. This pose-native resource connects sign language data with mainstream generative paradigms \cite{ma2024followposeposeguidedtexttovideo,zhang2025mimicmotionhighqualityhumanmotion,peebles2023scalablediffusionmodelstransformers}, making the definition of sign language motion representation and its associated evaluation protocol part of the scientific object itself rather than merely a preprocessing step \cite{Fang_2025_ICCV}. The resulting contribution combines a new data representation, task interface, and evaluation object with the scale of existing video-text corpora.

\noindent\textbf{Questions Addressed in This Paper.}~
Based on the above considerations, this paper does not aim to prove that SignVerse-2M is superior to existing resources for every sign language task. Instead, it focuses on three more specific questions. First, can large-scale public videos be stably transformed into a unified pose corpus suitable for sign language generation and directly compatible with modern pose-driven generation frameworks \cite{yang2023effective,cheng2025wan,ma2024followposeposeguidedtexttovideo}? Second, is this pose-space representation sufficient to support the training and comparison of multilingual generative models under real-world distributions \cite{yin2022mlslt,Fang_2025_ICCV,fang2026signxcontinuoussignrecognition}? Third, if used as an evaluation resource, what conclusions can it support, and where are its boundaries and limitations \cite{walsh2025slrtp2025signlanguageproduction,koller2020quantitative}? The remainder of the paper is organized around these three questions.

\section{Data Infrastructure and Pipeline}

\begin{figure*}[t]
    \centering
    \includegraphics[width=\textwidth]{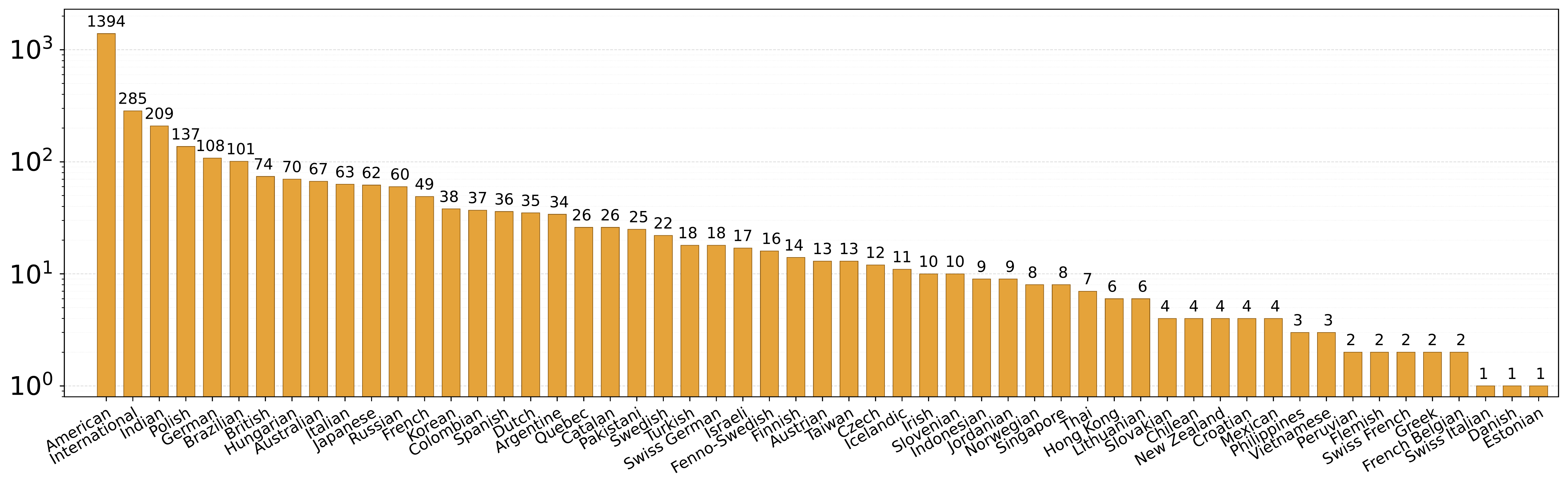}
    \caption{Distribution of content hours across sign languages in YouTube-SL-25. The x-axis lists sign language names, and the y-axis reports total hours on a logarithmic scale.}
    \label{fig:youtube_sl25_hours_logy}
\end{figure*}

\subsection{Data Source and Curation Pipeline}
\noindent\textbf{Data Source.}~
SignVerse-2M starts from publicly available multilingual sign language video resources and builds a large-scale corpus for pose-based modeling on top of them.
We maintain a manifest centered on \texttt{video\_id} and \texttt{sign\_language}, which drives the subsequent processing pipeline.
The raw metadata table in the current repository contains 39,196 video entries and roughly 2M timestamped subtitle segments; each video entry is fed into a traceable, resumable, and re-runnable processing pipeline with associated status records.
Accordingly, the focus of SignVerse-2M is to build an infrastructure for automatic sign language video processing that turns open-world videos into a unified, trainable, and publishable pose-native corpus.

\noindent\textbf{Pipeline Overview.}~
Figure~\ref{fig:signverse_data_pipeline} summarizes the overall construction path of SignVerse-2M.
The pipeline begins with manifest-driven sample management, proceeds through metadata and subtitle acquisition, and then moves into pose extraction, packaging, and publication.
This layout is useful because it makes clear that the corpus is not defined by a single preprocessing script, but by a sequence of recoverable and independently executable stages.

\noindent\textbf{Video Acquisition.}~
Before pose extraction begins, we first complete video-level data acquisition.
For each video ID in the manifest, the system retrieves the original platform metadata, downloads the raw video, and fetches the available subtitles in parallel.
This stage continuously writes back fields such as the title, duration, subtitle languages, raw video path, processing time, and error status, allowing the metadata table to serve as both a sample inventory and a runtime record.
As a result, the corpus construction process can proceed incrementally by entry and can resume from intermediate states after interruption.

\noindent\textbf{Subtitle Structuring and Language Signals.}~
Once subtitles are downloaded locally, they are converted into structured text objects instead of preserving the platform-exported format verbatim.
During processing, we remove WEBVTT control fields, timestamp lines, HTML tags, and zero-width characters, and apply rule-based normalization to repeated lines, broken punctuation, hyphen continuations, and ellipses.
At the output level, we preserve both segment-level subtitles with start and end times and compact document-level text intended for training-time consumption.
For English supervision, the system prioritizes native English subtitles; when no native English track exists, it automatically selects from available English translation variants and records the source language.
Each video therefore receives a structured caption JSON containing multilingual subtitle text, temporal boundaries, and the provenance of the English supervision.

\begin{figure*}[t]
    \centering
    \begin{minipage}[t]{0.465\textwidth}
        \vspace{0pt}
        \centering
        \includegraphics[width=\linewidth]{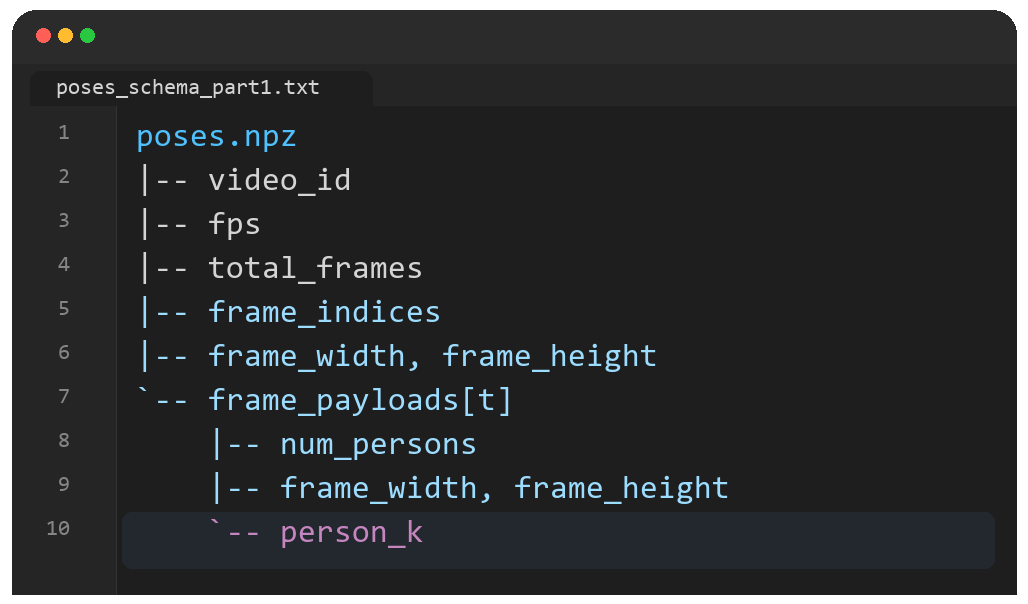}
    \end{minipage}
    \hfill
    \begin{minipage}[t]{0.50\textwidth}
        \vspace{0pt}
        \centering
        \includegraphics[width=\linewidth]{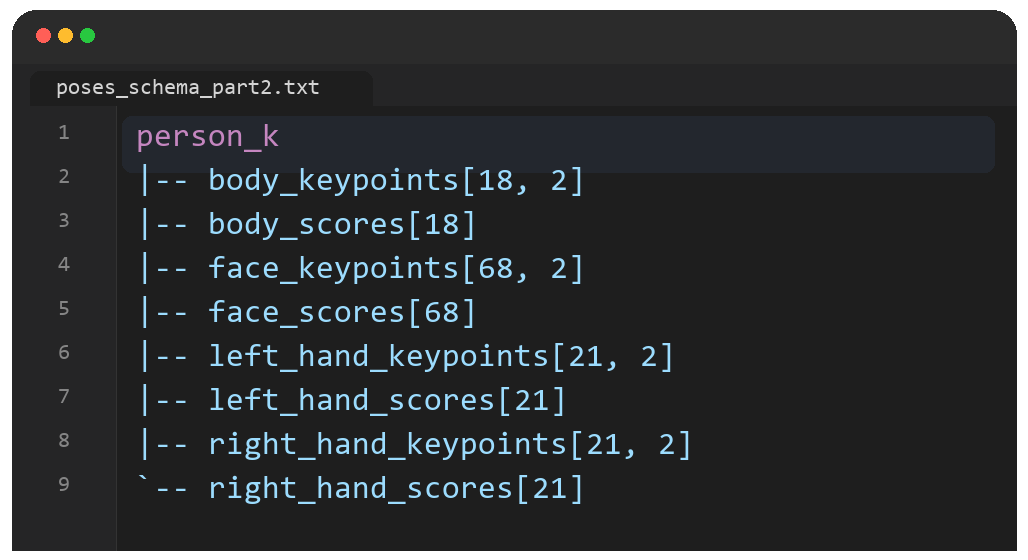}
    \end{minipage}
    \caption{\textbf{Schema of the released \texttt{poses.npz} payload.} The stored format is person-centric: each frame payload records the number of detected signers and organizes body, face, left-hand, and right-hand keypoints together with their confidence scores under each \texttt{person\_k}. This structure is the native representation released by SignVerse-2M and is the basis from which visualization scripts derive OpenPose-style aggregated renderings.}
    \label{fig:poses_schema}
\end{figure*}

\noindent\textbf{From Video-Text Data to Pose Data.}~
After videos and subtitles have been acquired, the system converts each video into a video-level pose sequence from which timestamp-aligned pose clips are indexed.
The current implementation decodes frames at 24 FPS and applies DWPose to extract body, hand, and facial keypoints from each frame.
To support large-scale processing, the pipeline can read frames directly from the \texttt{ffmpeg} stdout stream instead of first materializing all frames as JPG files; a legacy path is retained for debugging and fallback.
Pose outputs are aggregated per video into \texttt{poses.npz}, which stores the video ID, sampling rate, total frame count, frame indices, frame dimensions, and per-frame keypoint payloads.
After each video is processed, the system also writes a \texttt{.complete} marker and corresponding status records to distinguish unprocessed, running, and completed samples.
Through this stage, open-world sign language videos are reorganized into a unified pose representation and moved into the downstream training and publishing workflow.

\noindent\textbf{Processing Scale.}~
Converting the full corpus into this pose-native format required more than 80 GPUs over approximately five months, totaling about 288,000 GPU-hours for frame decoding, DWPose extraction, serialization, and status tracking. This computation reflects the one-time corpus construction cost; downstream users can work directly with the released pose sequences without repeating full-corpus extraction.

\section{The SignVerse-2M Corpus}

\subsection{Corpus Characteristics and Research Value}
\noindent\textbf{Data Organization and Corpus Value.}~
SignVerse-2M preserves the complexity of open-world data.
These videos come from different cameras, framing conditions, subtitle habits, and content genres, so the input conditions are inherently heterogeneous.
As a result, the corpus construction process must prioritize stable throughput, error recovery, and traceable runtime states on heterogeneous inputs rather than assuming a clean and controlled setting.
For downstream modeling, the resulting corpus is also closer to the data distribution that real deployments are likely to encounter.
At the organizational level, SignVerse-2M adopts a per-video directory structure.
Each \texttt{video\_id} corresponds to an individual directory containing subtitle files, a structured caption JSON, the pose result \texttt{poses.npz}, and a \texttt{.complete} marker indicating successful processing.
At the global level, the system also maintains \texttt{stats.npz} as a lightweight state index that records the processing status of each sample across metadata, subtitle, download, process, and upload stages.
This design enables both local recomputation at the level of individual videos and corpus-scale progress tracking and failure recovery.
The current manifest covers roughly 39K source videos and about 2M timestamped subtitle segments.
Each segment defines one pose clip by indexing its temporal interval in the corresponding video-level \texttt{poses.npz}; storing the full sequence once avoids duplicating frames while exposing segment-level training samples.
The predefined partitions are constructed at the source-video level, so every segment from one video remains in the same split.
An audit of the processed corpus records an empty-pose-frame rate of 1.89\%, with mean DWPose hand and face confidence scores of 0.653 and 0.967, respectively.
Together with the per-stage status records and visualization tools, these statistics support corpus inspection, while long-running execution, batch processing, and incremental updates keep construction reliable at scale.
From the repository implementation, downloading, pose extraction, and uploading are all separated into distinct stages and can be executed on selected subsets of samples, which leaves room for future expansion and maintenance.

\noindent\textbf{Cross-Lingual Scale.}~
Figure~\ref{fig:youtube_sl25_hours_logy} gives a direct view of the language distribution inherited from YouTube-SL-25.
The logarithmic y-axis makes the long-tail structure visible: a small number of languages contribute a large fraction of the total hours, while many other languages remain present at much smaller scales.
For SignVerse-2M, this distribution is not a side detail but a core property of the corpus, because it shapes multilingual training, transfer behavior, and the practical difficulty of benchmarking across languages.

\begin{table*}[t]
\vspace{-6pt}
\centering
\resizebox{1\linewidth}{!}{%
\begin{tabular}{@{}p{3.2cm}c|ccccc|ccccc@{}}
\toprule
  & & \multicolumn{5}{c}{DEV SET} & \multicolumn{5}{c}{TEST SET} \\
\multicolumn{1}{c|}{Model} & \multicolumn{1}{c|}{Language} & BLEU-4 & BLEU-3 & BLEU-2 & BLEU-1 & ROUGE & BLEU-4 & BLEU-3 & BLEU-2 & BLEU-1 & ROUGE \\ \midrule
\multicolumn{1}{r|}{\textbf{SignDW Transformer (40M)}} & ASL  & 11.27 & 17.11 & 25.82 & 34.97 & 35.05 & 9.65 & 15.51 & 24.40 & 33.85 & 33.71 \\
\multicolumn{1}{r|}{\textbf{SignDW Transformer (1.2B)}} & ASL  & 11.72 & 17.56 & 26.12 & 35.82 & 35.89 & 10.10 & 15.96 & 24.70 & 34.30 & 34.26 \\
\midrule
\multicolumn{1}{r|}{\textbf{SignDW Transformer (40M)}} & BSL  & 9.58 & 15.32 & 24.05 & 33.80 & 33.42 & 8.35 & 14.04 & 22.72 & 32.58 & 32.08 \\
\multicolumn{1}{r|}{\textbf{SignDW Transformer (1.2B)}} & BSL  & 10.06 & 16.06 & 25.07 & 35.02 & 34.77 & 8.80 & 14.49 & 23.12 & 33.23 & 33.03 \\
\midrule
\multicolumn{1}{r|}{\textbf{SignDW Transformer (40M)}} & GSL  & 9.82 & 15.68 & 24.48 & 34.23 & 33.98 & 9.69 & 15.48 & 24.29 & 33.82 & 33.58 \\
\multicolumn{1}{r|}{\textbf{SignDW Transformer (1.2B)}} & GSL  & 10.70 & 16.80 & 25.94 & 35.73 & 35.55 & 10.14 & 16.19 & 25.28 & 34.97 & 35.04 \\
\midrule
\multicolumn{1}{r|}{\textbf{SignDW Transformer (40M)}} & DSGS & 10.16 & 16.07 & 25.02 & 34.73 & 34.25 & 9.12 & 14.96 & 23.77 & 33.10 & 33.09 \\
\multicolumn{1}{r|}{\textbf{SignDW Transformer (1.2B)}} & DSGS & 10.70 & 16.70 & 25.61 & 35.49 & 35.56 & 9.57 & 15.41 & 24.07 & 33.73 & 33.87 \\
\bottomrule
\end{tabular}%
}
\caption{
\textbf{Text-to-pose baseline results on four sign languages.} We report results for \textbf{SignDW Transformer (40M)} and \textbf{SignDW Transformer (1.2B)} on ASL, BSL, GSL, and DSGS.
}\label{tab:mslp}
\end{table*}

\subsection{DWPose and Visualization}
\noindent\textbf{DWPose Representation and Visualization.}~
SignVerse-2M uses DWPose as the backend for pose extraction.
For each frame, the system extracts body, hand, and facial keypoints together with their confidence scores.
In the aggregated format, \texttt{poses.npz} stores not only video-level metadata but also per-frame payloads; each payload records the number of people, frame dimensions, and person-wise body, face, left-hand, and right-hand keypoints and scores.
This representation allows downstream methods to consume temporal inputs directly at the keypoint level without re-decoding the original videos.
The dataset repository also provides visualization scripts that reconstruct \texttt{poses.npz} into inspectable skeletal renderings.
Specifically, the scripts can read either aggregated or per-frame NPZ outputs and generate multiple visualization styles, including control-style skeletons, OpenPose-style renderings, and previews overlaid on the original video.
The underlying drawing functions render body, hand, and face keypoints separately while using confidence scores to modulate the display.
As shown in Figure~\ref{fig:poses_schema}, the released \texttt{poses.npz} format stores pose data in a person-centric structure rather than as a single flat OpenPose-style array.
These visualization interfaces are not intended as part of the training input itself; instead, they serve as inspection and debugging tools for quickly identifying missing keypoints, multi-person interference, local jitter, and abnormal frames.

\noindent\textbf{Processing Skeleton.}~
The released \texttt{poses.npz} format is organized as a person-centric pose payload.
This organization makes explicit that SignVerse-2M stores pose data in a person-centric DWPose-style payload, with body, face, and both hands grouped under each detected signer at each frame. The accompanying visualization scripts can reorganize the same information into an OpenPose-style aggregated representation for rendering, but the released corpus itself keeps the per-person structure because it is easier to inspect, serialize, and reuse in downstream modeling.

\section{Evaluation \& Baseline}
\begin{figure*}[t]
    \centering
    \includegraphics[width=0.97\textwidth]{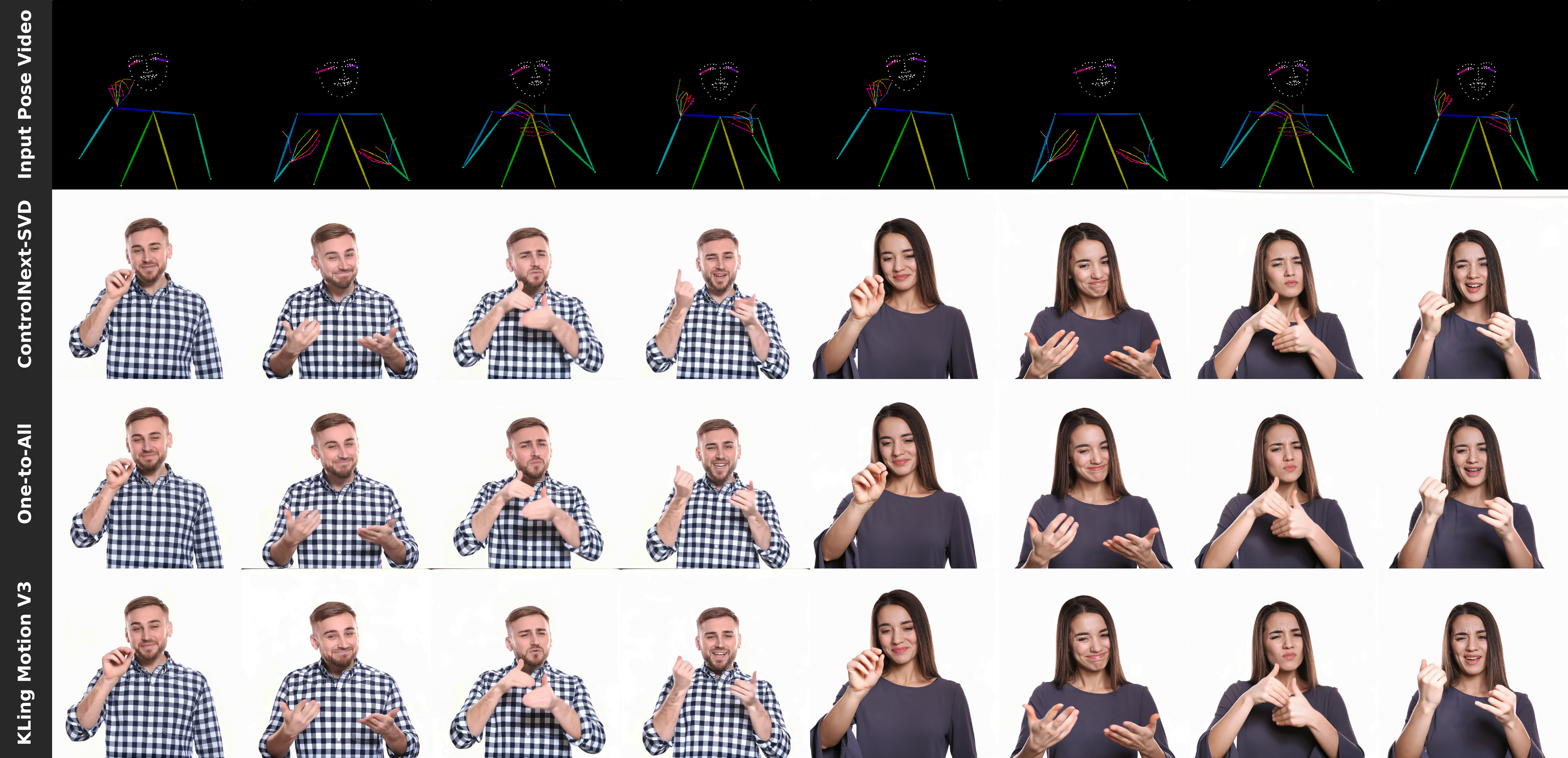}
    \caption{
    \textbf{Qualitative comparison of pose-conditioned sign video rendering.} The first row shows the input DWPose video. To remain compatible with model-specific preprocessing pipelines, we prepend a body-shape mask before feeding the pose video into each method. The next three rows compare \textbf{ControlNext-SVD}~\cite{peng2025controlnextpowerfulefficientcontrol}, \textbf{One-to-All}~\cite{shi2025onetoallanimationalignmentfreecharacter}, and \textbf{KLing Motion V3}~\cite{klingteam2025klingomnitechnicalreport} under the same pose sequence. This figure is intended to illustrate interface compatibility and coarse motion preservation rather than to compare the renderers themselves. For each rendering row, the left half shows the male signer and the right half shows the female signer.
    }
    \label{fig:4group}
\end{figure*}

Given the constraints of space and computational resources, we focus in this paper on a sign language generation baseline, as this setting most directly validates the practical usability of the proposed pose-native interface.

\subsection{Setup}

\noindent\textbf{Task.} Sign language generation aims to synthesize sign language motion from complex textual inputs. In this paper, our primary quantitative evaluation is carried out in pose space: First, the generated DWPose sequence is translated back into spoken-language text via SignX \cite{fang2026signxcontinuoussignrecognition}, and the recovered sentence is then compared with the original input. This back-translation protocol provides a broad proxy for semantic fidelity while avoiding additional variation introduced by downstream video rendering \cite{saunders2020progressive}. Second, we further apply the same back-translation idea after the pose-to-video step, using rendered videos as an additional renderer-level check of semantic preservation.
Third, we inspect the rendered videos qualitatively. These rendered-video evaluations are therefore used as compatibility and semantic-preservation checks rather than as a complete end-to-end assessment of sign video quality.



\noindent\textbf{Metrics.} We report BLEU-$n$~\cite{bleu} for n-gram overlap between the recovered and reference text, ROUGE~\cite{lin-acl-2004-rouge} for complementary text-overlap coverage, and DTW-MJE~\cite{berndt1994dtw,walsh2025slrtp2025signlanguageproduction} for mean joint-position error after temporal alignment of generated and reference poses. BLEU and ROUGE are higher-is-better, whereas DTW-MJE is lower-is-better.

\subsection{Results.}

Table~\ref{tab:mslp} summarizes the text-to-pose results on four representative sign languages: ASL, BSL, GSL, and DSGS. For all runs, we train the baseline for 200K steps with batch size 32 and an initial learning rate of 0.001 under the same multilingual DWPose interface, so that differences across rows mainly reflect model scale and language-specific data conditions rather than changes in the optimization setting. Overall, the proposed baseline achieves stable performance across all four settings, suggesting that the unified DWPose representation is usable for multilingual text-to-pose modeling under a shared training and evaluation interface. Across languages, the larger \textbf{SignDW Transformer (1.2B)} model is consistently stronger than the \textbf{SignDW Transformer (40M)} model, with the clearest gains appearing in BLEU-4 and BLEU-3, indicating better preservation of longer local motion patterns after back-translation. At the same time, the gap between languages remains non-negligible: ASL and GSL are relatively stronger, while BSL and DSGS remain more challenging, especially on the test split. We view this pattern as consistent with the multilingual and open-world nature of SignVerse-2M. The benchmark therefore serves primarily as a proof of feasibility and as an initial reference point for how model scale and language-specific data conditions affect pose-native sign generation.

\begin{table}[t]
\centering
\small
\setlength{\tabcolsep}{8pt}
\begin{tabular}{@{}lcccc@{}}
\toprule
Language & ASL & BSL & GSL & DSGS \\
\midrule
DTW-MJE $\downarrow$ & 4.68 & 5.21 & \textbf{4.59} & 6.27 \\
\bottomrule
\end{tabular}
\caption{\textbf{Motion-space evaluation across four sign languages.} DTW-MJE measures joint error after temporal alignment and is independent of the SignX back-translation evaluator.}
\label{tab:dtw_mje}
\end{table}

\begin{table}[t]
\centering
\small
\setlength{\tabcolsep}{3.2pt}
\resizebox{\columnwidth}{!}{%
\begin{tabular}{@{}lcccccc@{}}
\toprule
Model & B-4 $\uparrow$ & B-3 $\uparrow$ & B-2 $\uparrow$ & B-1 $\uparrow$ & R $\uparrow$ & DTW $\downarrow$ \\
\midrule
Progressive Transformers~\cite{saunders2020progressive} & 8.55 & 12.18 & 19.69 & 27.10 & 29.44 & 5.98 \\
SignDW Transformer (1.2B) & \textbf{10.10} & \textbf{15.96} & \textbf{24.70} & \textbf{34.30} & \textbf{34.26} & \textbf{4.68} \\
\bottomrule
\end{tabular}%
}
\caption{\textbf{Cross-architecture evaluation on ASL.} Both models use the same SignVerse-2M split and evaluation protocol. B-$n$, R, and DTW denote BLEU-$n$, ROUGE, and DTW-MJE.}
\label{tab:cross_architecture}
\end{table}

Table~\ref{tab:dtw_mje} complements back-translation with DTW-MJE, an evaluator-independent motion-space measure. Table~\ref{tab:cross_architecture} further applies the same ASL split and evaluation protocol to Progressive Transformers, showing that an independently developed text-to-pose architecture can directly use the SignVerse-2M representation.

Figure~\ref{fig:4group} provides a complementary qualitative view of this result. Beyond back-translation scores, the examples show that the DWPose-based representation used in SignVerse-2M can be interfaced directly with several modern pose-conditioned rendering systems, including both commercial and open-source pipelines, while preserving the signer's coarse motion structure and temporal progression. 
\begin{table}[t]
\centering
\small
\setlength{\tabcolsep}{5pt}
{%
\begin{tabular}{@{}p{2.9cm}cc@{}}
\toprule
\multicolumn{1}{c}{Renderer}  & \multicolumn{1}{c}{BLEU-4 $\uparrow$} & \multicolumn{1}{c}{ROUGE $\uparrow$} \\ \midrule
\multicolumn{1}{r|}{ControlNext-SVD~\cite{peng2025controlnextpowerfulefficientcontrol}} & 8.60 & 30.44 \\
\multicolumn{1}{r|}{One-to-All~\cite{shi2025onetoallanimationalignmentfreecharacter}} & 9.28 & 32.83 \\
\multicolumn{1}{r|}{KLing Motion V3~\cite{klingteam2025klingomnitechnicalreport}} & 10.01 & 34.65 \\
\bottomrule
\end{tabular}%
}
\caption{\textbf{Back-translation results after rendering.} Generated pose videos are rendered by the three systems used in Figure~\ref{fig:4group}, and the rendered outputs are evaluated by back-translation.}
\label{tab:prompt}
\end{table}
This point is important for our dataset motivation: earlier sign-language pipelines often relied on representations such as MediaPipe pose, OpenPose-style subsets, or SMPL-derived intermediate formats, which are useful for analysis but are not naturally aligned with the control interfaces adopted by recent image-to-video and talking-human generation models. In contrast, the representation used here provides a more direct connection to such systems, which is precisely the kind of systems-level compatibility that a pose-native benchmark is intended to provide.

Table~\ref{tab:prompt} provides a renderer-level view of this compatibility check by reporting back-translation scores after rendering the generated pose sequences into videos. 
Under this protocol, KLing Motion V3 yields the strongest recovered-text scores, followed by One-to-All and ControlNext-SVD, which is consistent with the qualitative impression in Figure~\ref{fig:4group} that better motion preservation in the rendering stage leads to better semantic recovery.

\section{Conclusion}
We present SignVerse-2M, a large-scale multilingual pose-native dataset for sign language research. Unlike existing sign language resources that are primarily derived from laboratory settings, the core contribution of SignVerse-2M is neither simply to scale up the amount of video data nor to introduce a new pose estimation algorithm. Instead, it systematically converts publicly available multilingual sign language videos from open environments into a unified DWPose representation, and builds around this representation a reusable data interface and evaluation foundation for sign language tasks. Through this resource, we aim to shift the focus of sign language generation research away from heterogeneous RGB video preprocessing and incomparable experimental settings toward motion representation, cross-lingual reuse, and evaluation itself.




% ...

\section*{LLM Usage Statement}
OpenAI Codex (GPT-5.6 and GPT-6) and various Anthropic Claude models were used to assist with design discussions, code development, and manuscript polishing. The core research idea originated entirely from the human authors, who remain responsible for and verified all technical content and results. All figures were manually designed and drawn by the authors without generative AI.

{
    \small
    \bibliographystyle{ieeenat_fullname}
    \bibliography{ref/main,ref/llm,ref/how2sign,ref/meta,ref/sds,ref/slt,ref/rl}
}

\clearpage
\setcounter{page}{1}
\maketitlesupplementary

\section{Implementation Details}

\subsection{Input and Output Representation}

SignDW Transformer is intentionally designed as a conventional text-to-pose baseline, allowing the experiments to evaluate the SignVerse-2M representation without conflating its effect with a specialized generation architecture. The input subtitle is tokenized with a vocabulary constructed from the corresponding training partition and encoded together with beginning- and end-of-sentence tokens.

Each target frame is represented by a 384-dimensional vector containing the coordinates and confidence scores of 18 body, 21 left-hand, 21 right-hand, and 68 facial keypoints. Coordinates are mapped into the same centered-square space used by the rendering interface. A scalar temporal counter is appended to obtain a 385-dimensional decoder input. For a sequence of $T$ valid frames, the counter at frame $t$ is $t/T$. Text and pose sequences are truncated or uniformly sampled to 96 positions, respectively; shorter sequences are padded and excluded from the objective by validity masks.

\subsection{Architecture}

The model follows an encoder--decoder Transformer design. Learned token embeddings and sinusoidal positional encodings feed the text encoder. The pose decoder projects each pose-plus-counter vector into the hidden space, adds a sinusoidal temporal encoding, and attends to the encoded text. Causal self-attention restricts each output frame to preceding pose frames. A linear prediction head recovers the 384-dimensional pose representation, while an auxiliary head predicts the normalized sequence length from the pooled text memory.

During training, the decoder sequence is shifted right: the first input is a zero pose and every subsequent position receives the preceding reference pose. This preserves teacher forcing without exposing the current target frame. Inference begins from the same zero pose and proceeds autoregressively. The 40M and 1.2B variants share the representation, decoding procedure, prediction heads, and training objective, differing only in model capacity.

\subsection{Objective and Optimization}

Let $\hat{\mathbf{p}}_{t}$ and $\mathbf{p}_{t}$ denote the predicted and reference 384-dimensional poses, and let $m_t$ indicate a valid, non-padding frame. The pose reconstruction objective is
\begin{equation}
\mathcal{L}_{\mathrm{pose}}=
\frac{\sum_{t}m_t\lVert\hat{\mathbf{p}}_{t}-\mathbf{p}_{t}\rVert_2^2}
{384\sum_t m_t}.
\end{equation}
The auxiliary length head is trained with mean squared error,
\begin{equation}
\mathcal{L}=\mathcal{L}_{\mathrm{pose}}+
\lambda_{\mathrm{len}}\mathcal{L}_{\mathrm{len}}.
\end{equation}
All reported language--model-size configurations are trained for 200K optimization steps with batch size 32 and initial learning rate $10^{-3}$. We use AdamW with cosine learning-rate decay. Samples are partitioned by source video before training, so subtitle intervals originating from one video cannot appear in both training and held-out partitions.

\section{Evaluation Protocol}

\subsection{Pose-Space Back-Translation}

The primary semantic evaluation follows the back-translation protocol used in sign language production research. The generated pose sequence is passed to the translation component of SignX~\cite{fang2026signxcontinuoussignrecognition}, producing a spoken-language sentence that is compared with the source text. For the four evaluated settings, the SignX translation component is trained for the corresponding language subset rather than applying a checkpoint trained for a different language or data distribution. The generator and evaluator use the same source-video-level partition definition, and reported scores are computed on held-out videos.

We report BLEU-1 through BLEU-4~\cite{bleu} and ROUGE~\cite{lin-acl-2004-rouge}. BLEU measures increasingly strict local $n$-gram agreement, whereas ROUGE provides a complementary coverage-oriented comparison. Pose-space back-translation is applied directly to generated DWPose sequences. Renderer-level back-translation applies the evaluator after the generated poses have passed through a pose-conditioned video renderer; this second setting therefore measures semantic information retained by the combined pose-generation and rendering path.

\subsection{DTW-MJE}

DTW-MJE provides a motion-space measurement that does not use the SignX text predictions. Generated and reference sequences are expressed in the same DWPose coordinate system. For generated frame $i$ and reference frame $j$, the local cost is the mean Euclidean distance over their mutually valid keypoints:
\begin{equation}
d(i,j)=\frac{1}{|\mathcal{V}_{ij}|}
\sum_{k\in\mathcal{V}_{ij}}
\left\lVert\hat{\mathbf{x}}_{i,k}-\mathbf{x}_{j,k}\right\rVert_2,
\end{equation}
where $\mathcal{V}_{ij}$ contains keypoints valid in both frames. Dynamic time warping~\cite{berndt1994dtw} finds the monotonic alignment path $\pi$ with minimum accumulated cost. We report
\begin{equation}
\operatorname{DTW\mbox{-}MJE}=
\frac{1}{|\pi|}\sum_{(i,j)\in\pi}d(i,j).
\end{equation}
This temporal alignment prevents differences in signing rate from being treated as frame-by-frame spatial errors. It complements back-translation by measuring pose trajectory agreement directly, following the motion-space evaluation direction used in sign language production benchmarks~\cite{walsh2025slrtp2025signlanguageproduction}.

\subsection{Cross-Architecture and Rendering Checks}

Progressive Transformers~\cite{saunders2020progressive} is trained on the same ASL partition and evaluated with the same back-translation and DTW-MJE implementations as SignDW Transformer. This comparison tests whether the released representation is usable by an independently developed text-to-pose architecture. The renderer study serves a different purpose: ControlNext-SVD, One-to-All, and KLing Motion V3 receive the same pose sequence and reference-person inputs. These outputs are used to verify interface compatibility and coarse semantic preservation; the study is not intended to rank general-purpose video renderers.

\section{Release Scope and Data Governance}

\subsection{Released Artifact}

SignVerse-2M does not re-host or redistribute source RGB videos. The released artifact consists of derived DWPose coordinates, structured subtitle records, processing metadata, predefined source-video-level partitions, status records, and the accompanying processing and visualization code. Samples are indexed by source \texttt{video\_id}, which preserves traceability to the public source while avoiding redistribution of the underlying audiovisual content.

The CC BY-NC 4.0 terms attached to the dataset apply only to the derived pose annotations and associated original metadata produced by this project. They do not grant or replace rights in the source videos, subtitles, or other upstream material, which remain governed by their original licenses and platform terms.

\subsection{Exclusion and Data Minimization}

The release provides a creator-initiated exclusion channel. Verified requests are processed within the published handling window, and removals are recorded in a versioned exclusion log so that subsequent releases consistently exclude the affected source identifiers. Because the corpus is organized by \texttt{video\_id}, exclusion can be applied to the corresponding pose, subtitle, metadata, and split entries without searching through opaque sample identifiers.

We also provide a face-landmark-excluded variant for applications that do not require facial keypoints. It retains body and hand motion while reducing identity-bearing channels. The responsible-use policy identifies signer re-identification as an unintended use. Although pose coordinates remove RGB appearance, temporally ordered motion may still contain behavioral or anthropometric signals; identifying, profiling, or inferring sensitive attributes of individual signers therefore falls outside the intended use of SignVerse-2M.

\end{document}